\pdfoutput=1

\documentclass[11pt]{article}

\usepackage[]{acl}

\usepackage{times}
\usepackage{latexsym}
\usepackage{booktabs}
\usepackage{todonotes}

\usepackage[T1]{fontenc}

\usepackage{amsmath,amsfonts,bm}

\def\eqref#1{equation~\ref{#1}}

\def\1{\bm{1}}

\DeclareMathAlphabet{\mathsfit}{\encodingdefault}{\sfdefault}{m}{sl}
\SetMathAlphabet{\mathsfit}{bold}{\encodingdefault}{\sfdefault}{bx}{n}

\usepackage[utf8]{inputenc}

\usepackage{microtype}

\usepackage{xcolor,colortbl}
\usepackage{subcaption}

\definecolor{Gray}{gray}{0.90}

\newcolumntype{a}{>{\columncolor{Gray}}c}

\newcommand{\mr}[1]{}
\newcommand{\todomr}[1]{}
\newcommand{\todomikel}[1]{}
\newcommand{\mikel}[1]{}

\title{On the Role of Parallel Data in Cross-lingual Transfer Learning}

\author{Machel Reid\thanks{\ \ Work done while at the University of Tokyo} \\
	Google Research \\\
	{\texttt{machelreid@google.com}} \\\And
  Mikel Artetxe \\
  Meta AI\\
  \texttt{artetxe@meta.com}}

\begin{document}
\maketitle
\begin{abstract}
While prior work has established that the use of parallel data is conducive for cross-lingual learning, it is unclear if the improvements come from the data itself, or it is the modeling of parallel interactions that matters. Exploring this, we examine the usage of unsupervised machine translation to generate synthetic parallel data, and compare it to supervised machine translation and gold parallel data. We find that even model generated parallel data can be useful for downstream tasks, in both a general setting (continued pretraining) as well as the task-specific setting (translate-train), although our best results are still obtained using real parallel data. Our findings suggest that existing multilingual models do not exploit the full potential of monolingual data, and prompt the community to reconsider the traditional categorization of cross-lingual learning approaches.
\end{abstract}

\section{Introduction}

\begin{figure}[t]
	\begin{subfigure}{\linewidth}
		\includegraphics[clip,width=\linewidth]{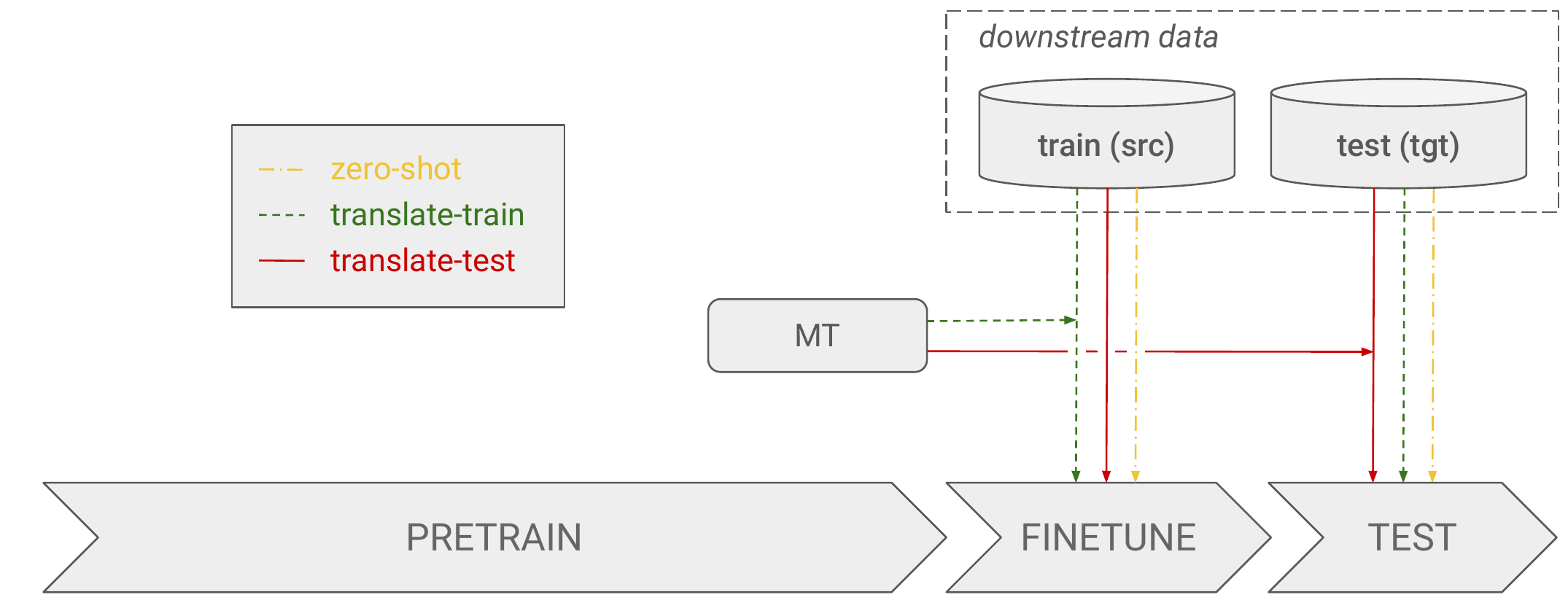}
		\caption{Traditional categorization}
		\label{subfig:traditional}
	\end{subfigure}
	\par\bigskip
	\begin{subfigure}{\linewidth}
		\includegraphics[clip,width=\linewidth]{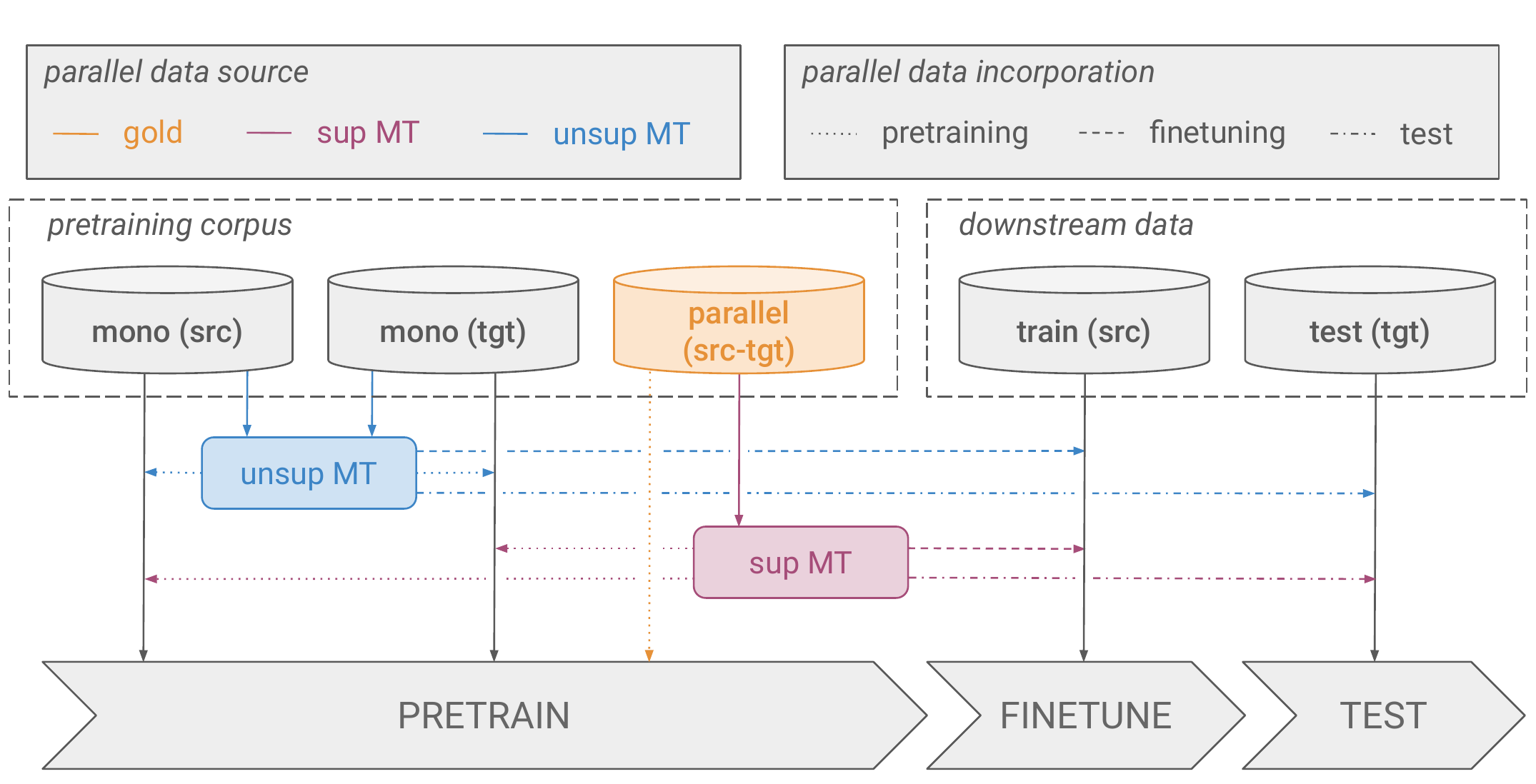}
		\caption{All possibilities to use different types of pretraining data}
		\label{subfig:ours}
	\end{subfigure}
	\caption{
		\textbf{Cross-lingual transfer settings.}
		Monolingual and parallel data can be used at different stages of the pipeline, either directly or indirectly through MT (b), but the traditional categorization falls short at capturing them (a).
	}
	\label{fig:summary}
\end{figure}

\textbf{Multilingual models} have been shown to generalize across languages in a zero-shot fashion \citep{pires-etal-2019-multilingual,conneau2019cross,conneau-etal-2020-unsupervised,kale2021mt5}. These models are usually pretrained on concatenated monolingual corpora in multiple languages using some form of language modeling or denoising objective. The models are then finetuned using labeled downstream data in the source language (typically English), which makes them capable of generalizing to the target language thanks to the aligned representations learned at pretraining.

While this paradigm does not require any data in the target language other than the monolingual pretraining corpus, prior work has reported improved results by incorporating \textbf{parallel data} into the pipeline, either at pretraining or finetuning time. During pretraining, parallel data has been incorporated through an auxiliary objective, such as Translation Language Modeling (TLM) in XLM \citep{conneau2019cross} or bitext denoising in PARADISE \citep{reid2021paradise}. Regarding finetuning, it is common to use Machine Translation (MT)---which is trained on parallel data under the hood---to translate the downstream training data into the target language(s) \citep{conneau-etal-2020-unsupervised}, which can be seen as a form of data augmentation.

Nevertheless, it is still unclear \textbf{why} parallel data is beneficial for cross-lingual transfer learning. Is the \textbf{data itself} that matters, given the additional information that it contains? Or is it the explicit \textbf{modeling} of parallel interactions that is important? To answer this question, we systematically compare the use of parallel data from different sources: ground truth parallel data, or synthetic parallel data generated by either supervised MT, unsupervised MT, or word-by-word translation. Most notably, our unsupervised MT variant relies on the exact same monolingual corpus used to pretrain the model, so any potential improvement can only come from the modeling side.

Our experiments on Natural Language Inference (NLI), Question Answering (QA) and Named Entity Recognition (NER) show that the explicit modeling of parallel interactions is indeed beneficial, demonstrating that existing pretraining and finetuning methods do not exploit the full potential of monolingual data. However, our best results are obtained using real parallel data---either directly or indirectly through supervised MT---showing that there is also some inherent value on it.

In the light of these results, we argue that the traditional categorization of cross-lingual transfer approaches into \textit{zero-shot}, \textit{translate-train} and \textit{translate-test} (Figure \ref{subfig:traditional}) falls short at capturing the required detail for a fair comparison across different approaches. %
Given this, we encourage further research on understanding what the contribution of monolingual and parallel data is, and how to best leverage them (directly or indirectly through MT, and at different parts of the pipeline), which requires thinking beyond the boundaries of the existing categorization (Figure \ref{subfig:ours}).

\begin{table*}[ht]
\centering
\resizebox{\textwidth}{!}{
\addtolength{\tabcolsep}{-3pt}
\begin{tabular}{llccccccalccccccalcccccccca}
\toprule
&& \multicolumn{7}{c}{XNLI (acc)}
&& \multicolumn{7}{c}{XQuAD (F1)}
&& \multicolumn{9}{c}{WikiANN (F1)}
\\
\cmidrule{3-9}
\cmidrule{11-17}
\cmidrule{19-27}
&
& en & ar & de & hi & fr & sw & \multicolumn{1}{c}{avg} &
& en & ar & hi & ru & th & vi & \multicolumn{1}{c}{avg} &
& en & ar & fr & hi & ru & th & vi & sw & \multicolumn{1}{c}{avg}
\\
\midrule
1) \ XLM-R &
& 83.9 & 71.9 & 75.2 & 69.1 & 77.4 & 62.2 & 73.3 & %
& 86.5 & 68.6 & 76.7 & 80.1 & 74.2 & 79.1 & 77.5 & %
& 81.3 & 53.0 & 80.5 & 73.0 & 69.1 & 1.3 & 79.4 & 70.5 & 63.5 %
\\
2) \textit{\quad + unsup MT} &
& 83.4 & 72.4 & 77.1 & 72.2 & 78.2 & 67.8 & 75.2 & %
& \bf 86.7 & 70.2 & 80.7 & 81.5 & 75.8 & 79.6 & 79.0 & %
& 81.3 & 54.1 & 82.1 & 74.9 & 71.1 & 3.8 & 80.7 & 71.7 & 64.9 %
\\
3) \textit{\quad + sup MT} &
& 83.2 & 74.4 & 77.5 & \bf 72.7 & 78.3 & 70.1 & 76.0 & %
& 86.6 & \bf 73.5 & 81.1 & \bf 83.0 & 77.4 & \bf 81.9 & \bf 80.7 & %
& 81.6 &  57.0 & 82.3 & 75.4 & 71.6 & \bf 5.8 & \bf 81.6 & 73.4 & 66.1 %
\\
4) \textit{\quad + gold} &
& \bf 84.0 & \bf 75.2 & \bf 77.7 & 72.4 & \bf 78.6 & \bf 70.4 & \bf 76.4 & %
& 86.3 & 72.3 & \bf 82.3 & 82.7 & \bf 78.2 & \bf 81.9 & 80.6 & %
& \bf 82.4 & \bf 57.3 & \bf 82.4 & \bf 75.6 & \bf 71.8 & 4.6 & 81.5 & \bf 73.7 & \bf 66.2 %
\\
\bottomrule
\end{tabular}
}
\caption{
\textbf{Pretraining incorporation results.}
We compare the original XLM-R model (1) with three variants where we continue pretraining it on either synthetic (2, 3) or real (4) parallel data. All models are finetuned on English downstream data and zero-shot transferred to the target language.
}
\label{tab:pretraining}
\end{table*}

\begin{table*}[ht]
\centering
\resizebox{\textwidth}{!}{
\addtolength{\tabcolsep}{-3pt}
\begin{tabular}{llccccccalccccccalcccccccca}
\toprule
&& \multicolumn{7}{c}{XNLI (acc)}
&& \multicolumn{7}{c}{XQuAD (F1)}
&& \multicolumn{9}{c}{WikiANN (F1)}
\\
\cmidrule{3-9}
\cmidrule{11-17}
\cmidrule{19-27}
&
& en & ar & de & hi & fr & sw & \multicolumn{1}{c}{avg} &
& en & ar & hi & ru & th & vi & \multicolumn{1}{c}{avg} &
& en & ar & fr & hi & ru & th & vi & sw & \multicolumn{1}{c}{avg}
\\
\midrule
1) \  XLM-R &
& 83.9 & 71.9 & 75.2 & 69.1 & 77.4 & 62.2 & 73.3 & %
& \bf 86.5 & 68.6 & 76.7 & 80.1 & 74.2 & 79.1 & 77.5 & %
& 81.3 & 53.0 & 80.5 & 73.0 & 69.1 & 1.3 & 79.4 & 70.5 & 63.5 %
\\
2) \textit{\quad + dict} &
& 83.7 & 72.6 & 77.6 & 70.7 & 78.9 & 65.6 & 74.9 & %
& -- & -- & -- & -- & -- & -- & -- & %
& -- & -- & -- & -- & -- & -- & -- & -- & -- %
\\
3) \textit{\quad + unsup MT} &
& 84.0 & 73.2 & 77.1 & 71.6 & 78.6 & 67.9 & 75.4 & %
& 86.0 & 70.4 & 80.3 & 81.0 & 76.3 & 79.8 & 78.9 & %
& 80.6 & 56.0 & 82.7 & 75.7 & 71.8 & 3.7 & 80.9 & 72.3 &  65.5 %
\\
4) \textit{\quad + sup MT} &
& \bf 84.2 & \bf 74.6 & \bf 78.2 & \bf 73.1 & \bf 79.4 & \bf 70.6 & \bf 76.7 & %
& 86.3 & \bf 73.2 & \bf 81.6 & \bf 83.4 & \bf 77.2 & \bf 81.4 & \bf 80.5 & %
& \bf 82.2 & \bf 57.4 & \bf 83.1 & \bf 76.4 & \bf 72.4 & \bf 5.2 & \bf 82.1 & \bf 73.4 & \bf 66.6 %
\\
\bottomrule
\end{tabular}
}
\caption{
\textbf{Finetuning incorporation results.}
We compare finetuning XLM-R on the original English data (1), and machine translated data through either word-by-word replacement (2), unsupervised MT (3) or supervised MT (4). %
}
\label{tab:finetuning}
\end{table*}

\section{Experimental setup}

\subsection{Tasks}

We run experiments on 3 tasks: NLI on XNLI \citep{conneau-etal-2018-xnli}, extractive QA on XQuAD \citep{artetxe-etal-2020-cross}, and NER on WikiANN \citep{pan-etal-2017-cross}. In all cases, we use the original training set in English, and evaluate transfer performance in other languages. Due to compute constraints, we restrict evaluation to the following set of languages: English (en), Arabic (ar), German (de), Hindi (hi), French (fr), Swahili (sw), Russian (ru), Thai (th) and Vietnamese (vi).

Our finetuning incorporation experiments in \S\ref{sec:finetuning} involve machine translating the training data into the target languages. For XNLI, we just translate the premise and hypothesis and leave the label unchanged. For XQuAD and WikiANN, which have token-level labels (as opposed to sequence-level), we translate the input text and project the answer spans by using the \texttt{awesome} \citep{dou2021word} word aligner , taking the aligned spans as the target labels.

\subsection{Model}

We use XLM-R base \citep{conneau-etal-2020-unsupervised} for all of our experiments, which was trained through Masked Language Modeling (MLM) on CC-100 (a monolingual corpus covering 100 languages). For finetuning, we experiment with learning rates of 1e-5, 5e-5, and 1e-4 using the Adam optimizer. We train for up to 10 epochs and choose the checkpoint with the best validation performance averaged across the languages in consideration.%

\subsection{Parallel data sources} \label{subsec:data}

We compare the following sources of parallel data in our experiments:

\paragraph{Gold.} Ground-truth parallel data generated by humans. We use the same parallel data as \citet{reid2021paradise}, which combines data from IWSLT, WMT, and other parallel corpora. %

\paragraph{Supervised MT.} Synthetic parallel data generated through a conventional MT system. The MT system is supervised, so this approach is also leveraging ground-truth parallel data indirectly. We use the 420M M2M-100 model \citep{m2m100}. %

\paragraph{Unsupervised MT.} Synthetic parallel data generated through an unsupervised MT system \citep{artetxe2018unsupervised,conneau2019cross}. The MT system is trained on a subset of the monolingual data used for pretraining, so this approach does not use any additional data neither directly nor indirectly, other than the synthetically generated one. More concretely, we use XLM-R base to initialize our unsupervised MT model, and finetune it in 16 directions (en$\leftrightarrow$\{ar,de,hi,fr,sw,ru,th,vi\} using the iterative denoising autoencoding and backtranslation approach proposed by \citet{conneau2019cross}.\footnote{\scriptsize{\url{https://github.com/facebookresearch/XLM}}} We train for a total of 750k iterations using a batch size of 128k tokens. We use 200MB of text from CC100 for each language, amounting to a total of 1.8GB of training data. %

\paragraph{Dictionary.} Synthetic parallel data generated through random word replacement with a dictionary. We use the same dictionaries as \citet{reid2021paradise}, which combine dictionaries from MUSE \citep{conneau2017word} and those extracted using word aligners \citep{Ostling2016efmaral}. Following \citet{reid2021paradise}, we replace words that are included in our dictionary with a probability of $0.4$.%

\section{Experiments and results}

\subsection{Pretraining incorporation} \label{sec:pretraining}

In these experiments, we incorporate parallel data into the pretraining process. We take XLM-R as our starting point, which was trained on monolingual data through MLM, and continue pretraining it on both MLM and TLM for 70k steps, using a batch size of 64k tokens. We use a learning rate of 5e-5 with a linear warmup and cosine decay schedule. We use the MLM objective 70\% of the time, and the TLM objective 30\% of the time. The latter applies the same masking objective over concatenated parallel sentences, and we compare different sources of parallel data as detailed in \S\ref{subsec:data}. For parallel data generated through MT, we translate a random subset of CC100 (keeping consistent with the data used in pretraining). The model is then finetuned on the downstream tasks using the original training data in English, and zero-shot transferred to the target languages.

We report our results in Table \ref{tab:pretraining}. We find that all variants incorporating parallel data outperform the original XLM-R model,\footnote{The skeptical reader might attribute this improvement to the additional training steps we perform, irrespective of the use of parallel data. However, we find strong evidence that the improvements are brought by the use of parallel data given that (i) XLM-R was trained until convergence using a huge amount of compute, and our continued training represents an insignificant fraction on top (96 GPU days, compared to 13k GPU days, or a relative 0.7\% further), and (ii) we get improvements in all target languages but not in English, suggesting that the additional steps improve the cross-lingual capabilities of the model but not its general quality.} and the improvements are consistent across all target languages. However, different from \citet{reid2021paradise}, we do not find any clear improvements on English. Regarding the source of parallel data, we find that supervised MT performs at par with gold data, even for less-resourced languages for which MT tends to suffer. Unsupervised MT lags behind them, but consistently outperforms the baseline.

These results suggests that the mere facilitation of parallel interaction is helpful even when not using any new data, but incorporating ground-truth parallel data brings further improvements. However, the way in which parallel data is incorporated---either directly or through MT---does not have any clear impact, as evidenced by the similar performance of supervised MT and gold.
\mikel{say something about the low performance in Thai? (see commented text)}

\subsection{Finetuning incorporation} \label{sec:finetuning}

In these experiments, we incorporate parallel data into the finetuning process. We translate the downstream training data in English into the rest of languages, and finetune XLM-R in the combined data in all languages. This is commonly referred to as \textit{translate-train-all} in the literature.

We report our results in Table \ref{tab:finetuning}. Similar to the finetuning incorporation, we find that incorporating parallel data outperforms the baseline in all tasks and target languages for all data sources that we explore. Supervised MT obtains the best results, followed by unsupervised MT and word-by-word translation with dictionaries. Similar to the pretraining incorporation results, this suggests that synthetic parallel data can bring improvements even when generated exclusively through monolingual data, but using real parallel data brings further improvements. Finally, we find that even simplistic ways to incorporate parallel signals can bring improvements, as evidenced by the dictionary replacement results.

\subsection{Discussion} \label{subsec:discussion}

While prior work has reported strong results from incorporating parallel data for cross-lingual transfer learning, our results show that this improvement can partly---but not exclusively---be attributed to the explicit use of a parallel training signal, which can also be achieved through unsupervised MT without the need for any real parallel data. In fact, we find that the facilitation of parallel interactions is more important than the use of real parallel data in all tasks but XQuAD, where the latter has a larger impact. Despite the popularity of multilingual pretrained models, which predominantly rely on monolingual data both for pretraining and finetuning, this calls into question the extent to which existing approaches are able to exploit the full potential of such monolingual data. %
In addition, it is striking that we obtain similar results for both pretraining and finetuning incorporation, as well as supervised MT and gold standard parallel data. While further evidence is necessary to draw a more definitive conclusion, this suggests that parallel data brings similar improvements regardless of when (pretraining vs. finetuning) and how (directly vs. indirectly through MT) it is incorporated.

\section{Reconsidering the categorization of cross-lingual learning approaches}

As illustrated in Figure \ref{subfig:traditional}, approaches to cross-lingual learning have traditionally been classified into 3 categories: \textit{zero-shot} (finetune a multilingual model on English and zero-shot transfer into the target language), \textit{translate-train} (translate the English training data into the target languages through MT and finetune a multilingual model), and \textit{translate-test} (translate the test set into English and run inference using a monolingual model). This distinction is primarily based on which stage of the pipeline MT is incorporated into. While relevant from a practical perspective, we believe that, if taken in a rigid manner, such a framework can hinder addressing the more fundamental question of what the contribution of each data source is, and how to best leverage each of them. \mr{and how to combine them; i.e. parallel data can be produced by the cross lingual model. Though perhaps this is implicility specified by ``leverage''}

More concretely, as shown in Figure \ref{subfig:ours}, there are different data \textbf{types} that one can use (monolingual source corpora, monolingual target corpora and parallel corpora, in addition to downstream data), which can be incorporated at different \textbf{stages} of the pipeline (pretraining, finetuning, testing) and via different \textbf{procedures} (directly or indirectly through MT). We argue that research in cross-lingual learning should aim to understand how the variants in each dimension as well the interactions between them impact downstream performance, which can require thinking beyond the boundaries of the 3 conventional categories. For instance, our variant using unsupervised MT to translate the downstream training data would fall within the definition of \textit{translate-train}. However, this approach is more comparable to \textit{zero-shot} in that it only uses monolingual data, and it would be unfair to compare it to conventional \textit{translate-train} systems that rely on parallel data to train the MT system. %

\section{Related work}

Prior work has explored the extent to which monolingual pretraining relies on knowledge transfer from unlabeled corpora by using synthetic data \citep{chiang2020pretraining,krishna-etal-2021-pretraining-summarization} or downstream data \citep{krishna2022downstream} instead, and similar ideas have also been explored in computer vision \citep{kataoka2020pretraining,asano2020critical}. However, to the best of our knowledge, we are first to examine if cross-lingual learning also relies on knowledge transfer from parallel data. Our use of synthetic parallel corpora is also connected with back-translation, which is widely used in MT \citep{sennrich-etal-2016-improving}. However, conventional MT systems are trained on parallel data, and back-translation is usually motivated as a way to leverage additional (monolingual) data. In contrast, our unsupervised MT variant does not use any additional data compared to regular pretraining.

\section{Conclusions}

In this work, we show that even model-generated parallel data can be useful for cross-lingual learning---greatly expanding the possibilities for multilingual models to improve their performance by taking advantage of their own machine translation capabilities. Given this, we advocate for investigating the optimal way to leverage monolingual and/or parallel data for cross-lingual learning, which might require thinking beyond the boundaries of the conventional \textit{zero-shot}, \textit{translate-train} and \textit{translate-test} categories.

\bibliography{anthology,custom}
\bibliographystyle{acl_natbib}

\appendix

\end{document}